\documentclass[10pt,twocolumn,letterpaper]{article}

\usepackage{wacv}
\usepackage{times}
\usepackage{epsfig}
\usepackage{graphicx}
\usepackage{amsmath}
\usepackage{amssymb}

\usepackage{mathtools}
\usepackage{algorithm,algpseudocode}
\usepackage{authblk}

\def\wacvPaperID{533} 

\wacvfinalcopy 

\ifwacvfinal
\def\assignedStartPage{1} 
\fi

\ifwacvfinal
\usepackage[breaklinks=true,bookmarks=false]{hyperref}
\else
\usepackage[pagebackref=true,breaklinks=true,colorlinks,bookmarks=false]{hyperref}
\fi

\ifwacvfinal
\else
\fi

\begin{document}

\title{MetaPix: Domain Transfer for Semantic Segmentation by Meta Pixel Weighting}


\author{Yiren Jian$^{1,}$\thanks {Both authors contributed equally to this research.}  \hspace{1em} Chongyang Gao$^{ 2,*}$ }
\affil{ \vspace{-0.5em} $^1$ Dartmouth College $^2$ Northwestern University
}
\affil{
\vspace{-0.5em} \tt\small yiren.jian.gr@dartmouth.edu \hspace{0.2em} chongyanggao2026@u.northwestern.edu}

\maketitle

\begin{abstract}
   Training a deep neural model for semantic segmentation requires collecting a large amount of pixel-level labeled data. To alleviate the data scarcity problem presented in the real world, one could utilize synthetic data whose label is easy to obtain. Previous work has shown that the performance of a semantic segmentation model can be improved by training jointly with real and synthetic examples with a proper weighting on the synthetic data. Such weighting was learned by a heuristic to maximize the similarity between synthetic and real examples. In our work, we instead learn a pixel-level weighting of the synthetic data by meta-learning, i.e., the learning of weighting should only be minimizing the loss on the target task. We achieve this by gradient-on-gradient technique to propagate the target loss back into the parameters of the weighting model. The experiments show that our method with only one single meta module can outperform a complicated combination of an adversarial feature alignment, a reconstruction loss, plus a hierarchical heuristic weighting at pixel, region and image levels.
\end{abstract}

\section{Introduction}

Deep learning has achieved state-of-the-art results in computer vision \cite{mask-rcnn, Pham_2021_CVPR}, language processing \cite{trasformer, devlin-etal-2019-bert}, protein modeling \cite{AlphaFold2021} and biomedical imaging \cite{Li:18}. However, this technique heavily relies on the large available labeled data for success. Semantic segmentation, which requires data to be labeled at the pixel level, presents an even tougher challenge for collecting labeled datasets, \ie, labeling a single $2048 \times 1024$ image may take up to one and a half hours \cite{Cordts2016Cityscapes}. Thus, the dataset like CityScapes \cite{Cordts2016Cityscapes} for street view understanding has only less than 3000 fine-grained labeled images. To combat the data scarcity problem, many methods utilize synthetic data from video games \cite{Richter_2016_ECCV} to expand the training dataset \cite{sun2019CVPR}. This synthetic data from video games comes with labels that are cheap to collect from the back-end physical engines. However, directly training a model with synthetic examples from the video games will suffer poor performance due to the domain gap, \eg, the inconsistency in textures, illuminations between synthetic and real-world data. Thus, minimizing this domain gap is the key to deploying a model trained under synthetic data to the real world. 

In this work, we follow the domain transfer learning (DTL) setting from a recent work \cite{sun2019CVPR}, to learn a model with an insufficient amount of real labeled examples and a large amount of labeled synthetic examples. Note that this is different from the widely studied unsupervised domain adaptation (UDA) \cite{zhang2021prototypical, vu2018advent, Tsai_adaptseg_2018, zou2018unsupervised, luo2019Taking, chang2019all, li2019bidirectional, Tsai_adaptseg_ICCV19, Chen_2019_ICCV, pan2020unsupervised} problem, which assumes labeled synthetic data and unlabeled real data in training. 

Adversarial training, distance minimization, or self-training are common techniques in UDA. While it is straightforward to adapt the latest developed methods from UDA into DTL, \eg, adding a supervision loss from the labeled target dataset to the existing UDA methods, the experiments from a prior work \cite{sun2019CVPR} and our new experiments show that this leads to only marginal improvements. 

Hierarchical Region Selection \cite{sun2019CVPR} is the first method explicitly designed to solve the problem of DTL in street view understanding. That work mitigates the domain gap by selecting part of pixels/regions from synthetic data to train the network. The selection modules are trained to predict weighting maps by a heuristic that the weighted synthetic data should look similar to the real data. In their implementation, this is achieved by minimizing the L2 norm between the weighted synthetic data and the real data. Though minimizing this L2 distance intuitively filters out pixels and regions in synthetic examples that are less similar to the real data, it has no guarantee or enforcement on minimizing the target loss on real data. To further improve the performance, Hierarchical Region Selection also incorporates a generator and a discriminator to devise a reconstruction loss and an adversarial loss on the output of the fully convolution network's encoder, which presents challenges in stabilizing the training and finding good balance between different losses.

\begin{figure*}[!t]
\centering
\includegraphics[width=\textwidth]{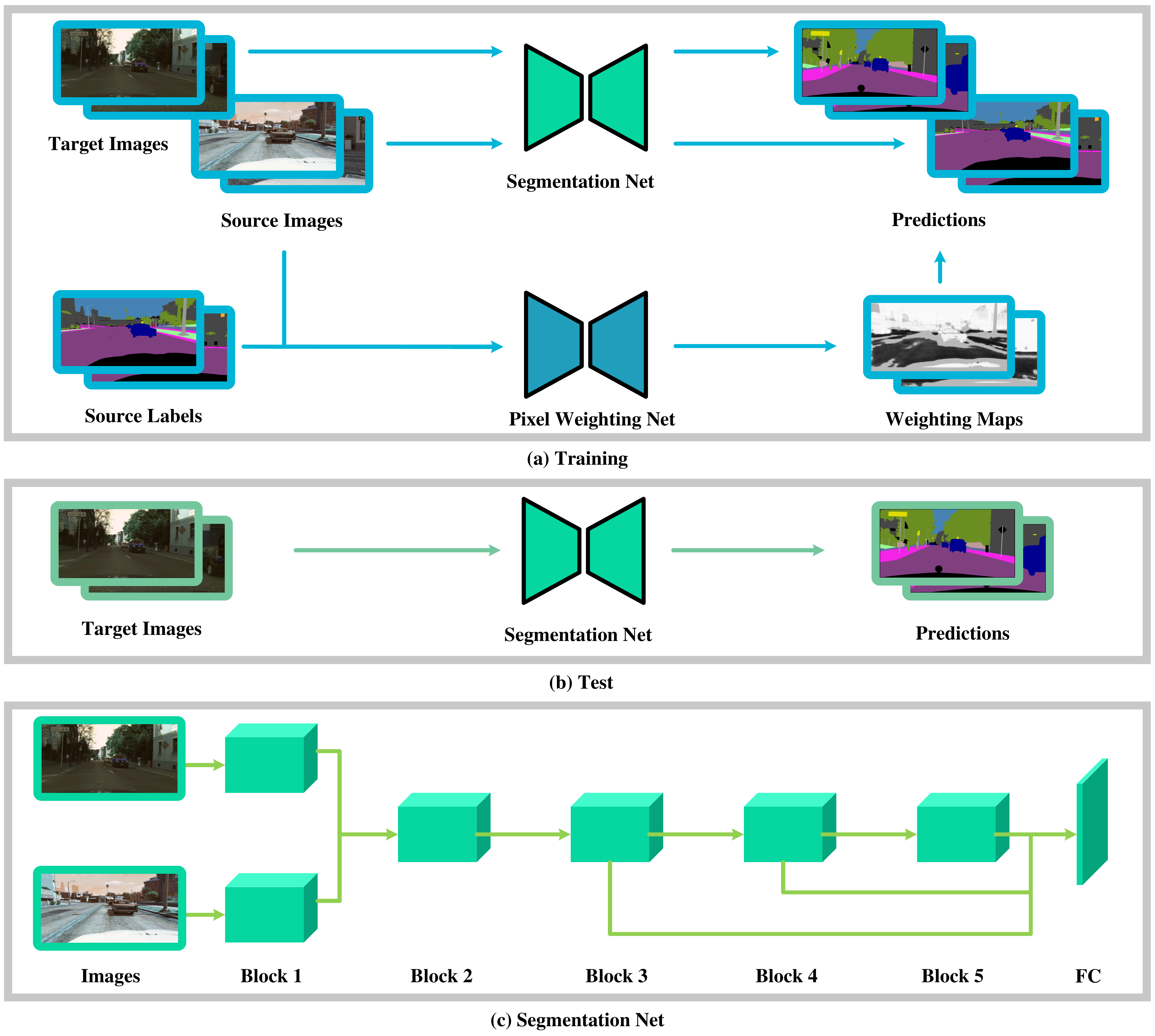}
\caption{Overview of our proposed method. (a) The segmentation network is trained with real images from the target domain and synthetic images from the source domain. While the loss computed from the real target image is the unweighted mean across all pixels as standard supervised training, a pixel-level weight scales the loss from the synthetic source image. This pixel-level weight is obtained by another weighting model with the synthetic source image as the input. The weighting model is meta-trained, which is further discussed in our method part. (b) During the testing, the pixel weighting network is discarded. (c) We empirically find that splitting the segmentation network at the first block with specific heads for the target domain and source domain is beneficial. This is further discussed in Ablations~\ref{sec:AB}.}
\label{fig:arch}
\end{figure*}

Here, we solve the DTL in the meta learning paradigm, \ie, the weighting model to select pixels should be learned in such a way that: learning with weighted synthetic data, the target loss, \ie, segmentation loss on real data will reduce. The idea of the meta learning is straightforward and does not require sophisticated techniques, \eg, adversarial learning and more complex losses. This meta learning of the weighting model is achieved by the gradient-on-gradient technique to propagate gradients from a target loss through the segmentation network to the weighting model. Actually, meta learning is only possible if we are provided with the labels in the target domain to construct that meta loss. This makes our method exclusively applicable to the setting of DTL but not UDA. Though gradient-based meta learning has shown promising in few-shot classification on small datasets, we prove its effectiveness in a large-scale semantic segmentation task with real-world application in street view understanding using the CityScapes dataset.  

Our method only learns jointly with meta-weighted synthetic examples and real examples without any sophisticated tricks like the adversarial alignment and complex losses from UDA, that are used by Hierarchical Region Selection. The improved experimental results of our method show that the superiority of our meta-learning based weighting module over the prior work.

\begin{figure*}[h!]
\centering
\includegraphics[width=\textwidth]{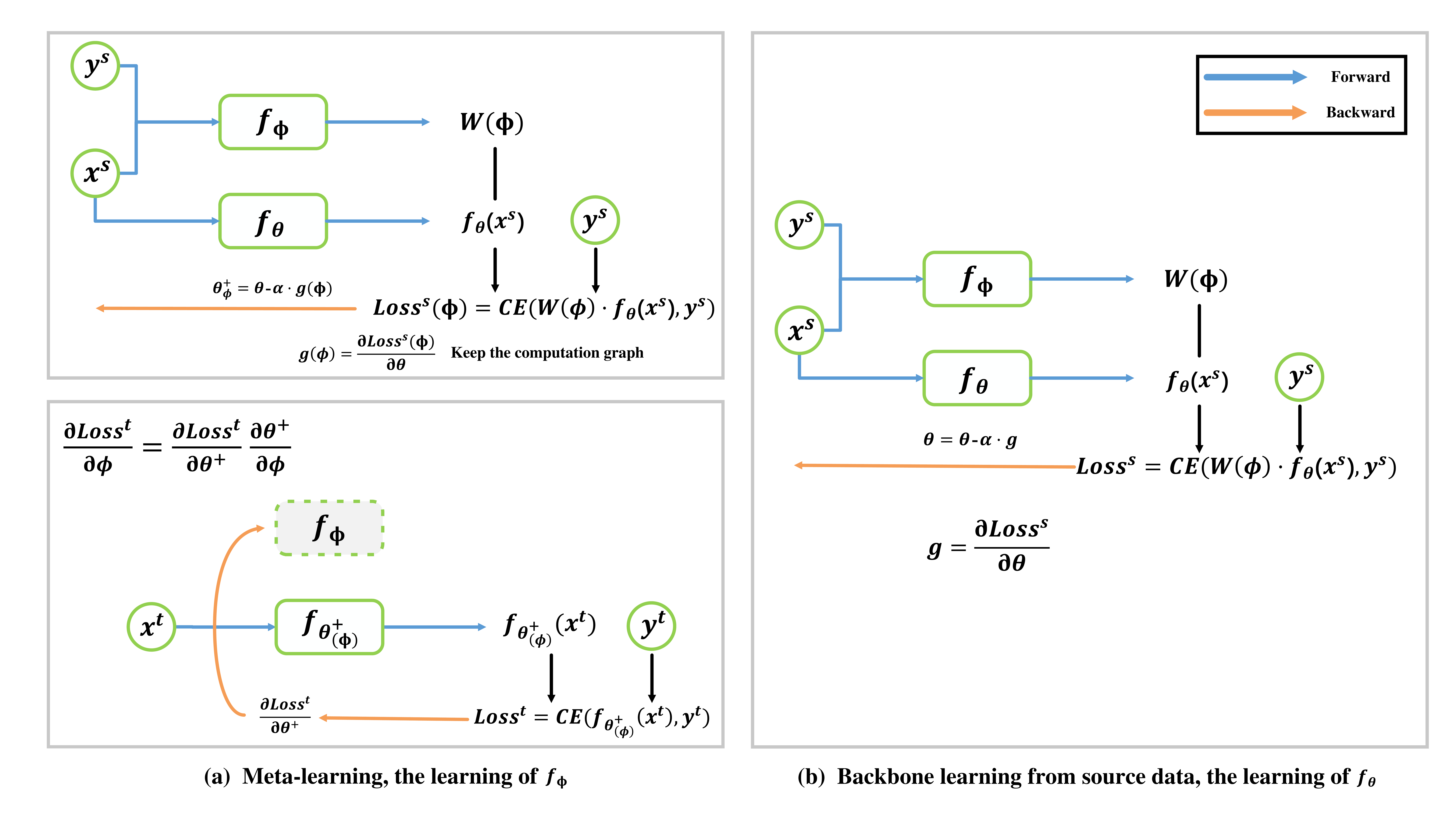}
\caption{Workflow of our proposed method in Section~\ref{sec:MLW} and Section~\ref{sec:LS}. The meta-learning of the weighting model $f_{\phi}$ composes of two forward-backward passes, one pass with source image $x^{s}$ and another pass with target image $x^{t}$. (upper part of figure(a)) It uses source images $x^s$ to compute a weighted source loss (create and retain computation graph) with both weighting model $f_{\phi}$ and the segmentation network $f_{\theta}$. The gradient on $\theta$ is computed from this source loss and it is used to update the segmentation network for one gradient step. Note that because we create the computational graph, the source loss $Loss^{s}(\phi)$, the gradient $g(\phi)$ and the updated parameters of the segmentation network $\theta^{+}(\phi)$ are all functions of $\phi$, which enables the following second forward-backward pass to learn $\phi$. (lower part of figure(a)) The target image $x^t$ and the updated segmentation network $\theta^{+}$ are used to compute a target loss $Loss^{t}$. This loss $Loss^{t}(\phi) $, as a function of $\phi$ can be used to update $\phi$, by gradient-on-gradient technique. (b) Once the weighting model $f_{\phi}$ is trained, it is used to compute the weight $W(\phi)$ from source image $x^s$. The final source loss is the weighted mean of cross-entropy between the prediction $f_{\theta}(x^s)$ and the label $y^s$. }
\label{fig:learning}
\end{figure*}

\section{Related Work}

\subsection{Semantic Segmentation} 
Semantic segmentation is an important task in computer vision and is crucial to many real-world applications, \eg, medical image processing, autonomous driving and retail applications. Semantic segmentation is a pixel-level classification task. One of the seminal works, the fully convolution network (FCN)~\cite{DBLP:conf/cvpr/LongSD15} is the first method using deep learning to solve this problem in an end-to-end manner. The state-of-the-art results on semantic segmentation benchmarks~\cite{DBLP:journals/ijcv/EveringhamEGWWZ15,Cordts2016Cityscapes,DBLP:conf/cvpr/RosSMVL16,DBLP:conf/cvpr/ZhouZPFB017,DBLP:conf/cvpr/CaesarUF18,Richter_2016_ECCV} are achieved by DeepLab series~\cite{DBLP:journals/corr/ChenPKMY14,DBLP:journals/pami/ChenPKMY18,DBLP:journals/corr/ChenPSA17}, which deal with multiscale context by Atrous Spatial Pyramid Pooling (ASPP). Besides, recent works utilize attention mechanisms~\cite{DBLP:conf/cvpr/FuLT0BFL19,DBLP:conf/iccv/HuangWHHW019,DBLP:conf/eccv/ZhaoZLSLLJ18,DBLP:journals/corr/abs-2005-10821,DBLP:journals/corr/abs-2004-08955,DBLP:conf/eccv/YuanCW20,DBLP:conf/cvpr/HouZCF20}, statistical analysis~\cite{zhu2021learning} and advanced pooling techniques~\cite{DBLP:conf/cvpr/HouZCF20}.

However, those models, with large capacities, require a large amount of labeled data to train. Ground truth labels in semantic segmentation are notoriously hard to obtain because one has to label every single pixel of images into a class. As an alternative, practitioners rely on auxiliary synthetic images to extend the training dataset~\cite{DBLP:conf/cvpr/RosSMVL16,Richter_2016_ECCV}. Those images are often collected from video games where pixel labels are easy to retrieve from the back-end physical engines. For example, GTA5 dataset~\cite{Richter_2016_ECCV} collects images from a video game that is built upon the street view of Los Angeles. The GTA5 dataset shares the label space with CityScapes, which allows joint training of a model on these two datasets. 

\subsection{Domain Adaptation} 
Domain adaptation is about transferring the knowledge of a model learned from one domain that is usually with plenty of labeled data to another. Domain adaptation is mostly divided into adversarial methods \cite{BSP_ICML_19, pmlr-v80-hoffman18a, Haoran_2020_ECCV, zhang2019domain}, distance-based methods \cite{Balaji_2019_ICCV, Li20DCAN, inproceedings, dcoral} and self-training methods \cite{zhang2020label, takahashi2020partially, pandey2020skin, cui2020hda}.

In the following, we mainly review domain adaptation methods for semantic segmentation. The most wildly studied setting of domain adaptation is unsupervised domain adaptation (UDA), where the source synthetic domain has the ground truth labels, but the target real domain does not. Inspired by cycleGAN \cite{CycleGAN2017}, CyCADA \cite{pmlr-v80-hoffman18a} enforces a cycle consistency and an adversarial alignment on the feature space. AdvEnt \cite{vu2018advent} minimizes the entropy of outputs between synthetic and real examples by adversarial training. MaxSquareLoss \cite{Chen_2019_ICCV} proposes a new loss to circumvent the unstable adversarial training commonly observed in UDA. 

Aligning of the two domains can also be regularized by content consistency \cite{DBLP:conf/eccv/LiKLWY20} and class consistency \cite{DBLP:conf/nips/ZhangZ0T19}. Other works attempt to minimizing the domain gap at input image space \cite{DBLP:conf/cvpr/0001S20, DBLP:conf/cvpr/YangCS20}, with auxiliary depth information \cite{DBLP:conf/iccv/VuJBCP19}, or curriculum learning \cite{DBLP:journals/pami/ZhangDFG20}. 

Besides aligning the input image spaces or feature spaces between the synthetic and the real domain, self-training has been a simple but effective approach. Self-training or pseudo-labeling \cite{zhang2021prototypical, li2019bidirectional, pan2020unsupervised}, first learns a model with the labeled dataset and uses the learned model to pseudo label the unlabeled set. Then the model is further trained with the extended dataset that contains both the labeled set and pseudo labeled set. Self-training can be improved by several iterations of refinements or selectively filtering low confident examples \cite{DBLP:conf/eccv/LiKLWY20}.

\subsection{Meta Learning} 
Meta learning algorithms are mostly designed for few-shot learning \cite{Ravi2017OptimizationAA, Sun_2019_CVPR}, fast model adaption \cite{pmlr-v70-finn17a} and the context of learning to optimize \cite{learn2learn}.

Gradient-based meta learning is first introduced by MAML \cite{pmlr-v70-finn17a} that aims at the fast adaptation of parameters of the network to the new tasks. Since the success of meta learning in solving few-shot learning tasks \cite{Flennerhag2020Meta-Learning, rusu2018metalearning, DBLP:conf/iclr/RaviL17}, meta learning has been applied to tasks in auxiliary learning, self-supervised \cite{liu2019maxl} learning, and semi-supervised learning. MAXL \cite{liu2019maxl} meta-learns a network to generate auxiliary tasks for improving the performance on the target task in a self-supervised manner. Meta Pseudo Label \cite{Pham_2021_CVPR} achieves the state-of-the-art on ImageNet benchmark by meta learning a model to pseudo label the 300M JFT dataset in a semi-supervised learning setting. Meta weighting of examples \cite{ren18l2rw} (on the image level) has also shown to be beneficial in reinforcement learning \cite{NIPS2019_8724}. 

While most meta learning methods are applied to small datasets in few-shot classification tasks or few-shot semantic segmentation \cite{Tian_Wu_Qi_Wang_Shi_Gao_2020}, we firstly apply this paradigm to domain transfer of semantic segmentation with real-world application in street view understanding and show its effectiveness. 


\section{Method}
In this section, we formally introduce our problem setting and the algorithm, which we name as \textit{Meta Pixel Weighting} (MetaPix). The procedures are shown in Algorithm~\ref{alg:a1} and Figure \ref{fig:learning}. 

\subsection{Problem Overview}
Assuming we have two labeled datasets, a source synthetic dataset $X_{s}=\{(x^{s}, y^{s})\}$ and a target real dataset $X_{t}=\{(x^{t}, y^{t})\}$, where $x^{s}, x^{t} \in \mathbb{R}^{H \times W \times 3}$. In Domain Transfer Learning \cite{sun2019CVPR}, we assume that the source synthetic dataset is much larger than the target real dataset, \ie, $|X_{s}| \gg |X_{t}|$. We care only about the performance of the model on the validation part of the target real dataset. 

We have two networks in our algorithm: a fully convolution network FCN as the segmentation network $f_{\theta}$ for semantic predictions, and another auxiliary weighting network $f_{\phi}$ used during the training. $f_{\theta}$ takes the images as input and outputs a tensor of size $[B, cls, H, W]$, where $B, cls, H, W$ are number of batches, number of label classes, the height of images and width of images. $f_{\phi}$ takes the concatenated input images and one-hot encoding of the labels as input and outputs a tensor of size $[B, 1, H, W]$ as the weighting map to weight source synthetic data. $f_{\theta}$ will be trained with pixel-weighted cross-entropy loss on source synthetic dataset $X_{s}$ and unweighted cross-entropy loss on target real dataset $X_{t}$. $f_{\phi}$ will be learned by meta learning using gradient-on-gradient.

Our weighting network $f_{\phi}$ is similar to the pixel-level selection module of $W^{1}$ in Hierarchical Region Selection \cite{sun2019CVPR}, but we learn this module by meta-learning. Besides, we do not have complex image-level and region-level weighting used in Hierarchical Region Selection.

\subsection{Meta Learning Weights for Source Examples}
\label{sec:MLW}
In this section, we learn the optimal weighting model $f_{\phi}$ given the current estimate of the segmentation model $f_{\theta}$. We first feed the segmentation model $f_{\theta}$ with source data $x^{s}$, and create the computational graph.
\begin{align}
    Loss^{s}(\phi) = \sum_{h,w} \frac{W_{h,w}(\phi)}{HW} CrossEntropy(f_{\theta}(x^{s}_{h,w}), y^{s}_{h,w})
\end{align}
where $W_{h,w}(\phi)$ is computed by a forward pass of concatenated source examples and source labels through the weight network, i.e., $W_{h,w} = f_{\phi}(concat(x^{s}_{h,w}, {y}^{s}_{h,w}))$. The loss on the source data is now a function of $\phi$. We take the derivative of $Loss^{s}(\phi)$ w.r.t parameters of segmentation model $f_{\theta}$, and retain the computation graph.
\begin{align}
    g_{\theta}(\phi) = \frac{\partial Loss^{s}(\phi)}{\partial \theta}
\end{align}
The parameters of the segmentation model $f_{\theta}$ is then updated by one gradient descent with learning rate $\alpha$.
\begin{align}
    \theta^{+}(\phi) = \theta - \alpha g_{\theta}(\phi)
\end{align}
$\theta^{+}(\phi)$ is now the parameters of the segmentation model. We enforce that parameters $\theta^{+}(\phi)$ learned from source should be optimized w.r.t the target task, \ie, we construct a target loss for $\theta^{+}(\phi)$, by a second forward pass on segmentation model $f_{\theta^{+}}$ with target images $x^{t}$.
\begin{align}
    Loss^{t}(\phi) = \sum_{h,w} \frac{1}{HW} CrossEntropy(f_{\theta^{+}(\phi)}(x^{t}_{h,w}), y^{t}_{h,w})
\end{align}
Then, using the retrained computational graph, we can compute the gradient for $\phi$ from this loss using gradient on gradient.
\begin{align}
    \frac{\partial Loss^{t}(\phi)}{\partial \phi} = \frac{\partial Loss^{t}(\phi)}{\partial \theta^{+}(\phi)} \frac{\partial \theta^{+}(\phi)}{\partial \phi}
\end{align}

In practice, in order to have a good estimation of segmentation model $f_{\theta}$ to learn the weighting network $f_{\phi}$, we first learn a segmentation model $f_{\theta}$ using $\{X^{t}, X^{s}\}$ jointly before we meta learning the weight network $f_{\phi}$.

\subsection{Learning the Segmentation Model with Weighted Source Loss}
\label{sec:LS}
Assuming we already have a learned (fixed) weighting network $f_{\phi}$, we can then learn segmentation model $f_{\theta}$ using both $X_{s}=\{(x^{s}, y^{s})\}$ and $X_{t}=\{(x^{t}, y^{t})\}$. When training with $X_{t}=\{(x^{t}, y^{t})\}$, the unweighted mean of cross entropy at each pixel is used for an end-to-end training.
\begin{align}
    Loss^{t} = \sum_{h,w} \frac{1}{HW} CrossEntropy(f_{\theta}(x^{t}_{h,w}), y^{t}_{h,w})
\end{align}
When training with $X_{s}=\{(x^{s}, y^{s})\}$, we first compute the weighting using $f_{\phi}$.
\begin{align}
    W_{h,w} = f_{\phi}(concat(x^{s}_{h,w}, {y}^{s}_{h,w}))
\end{align}
Then the loss for source data is computed by the weighted mean of cross entropy.
\begin{align}
    Loss^{s} = \sum_{h,w} \frac{W_{h,w}}{HW} CrossEntropy(f_{\theta}(x^{s}_{h,w}), y^{s}_{h,w})
\end{align}
The finally loss for training the segmentation model $f_{\theta}$ is the sum of $Loss^{s}$ and $Loss^{t}$.
\begin{align}
    Loss = Loss^{s} + Loss^{t}
\end{align}

\begin{algorithm}
\caption{MetaPix}\label{alg:a1}
\begin{algorithmic}[1]

\State \textbf{Input}: $X_{s}, X_{t}$ \Comment{Source domain, target domain} 
\State \qquad $N_{1}, N_{2}, N_{3}$ \Comment{Number of iterations}
\State \qquad $G$ \Comment{Number of generations}

\State \textbf{Initialization}: Weighting Network $\phi$, from scratch.
\State Segmentation network $\theta$, from ImageNet Pretrained.

\Procedure{:}{}
\Repeat
    \State $\{x^{s}, y^{s}\} \gets X_{s}$
    \State $\{x^{t}, y^{t}\} \gets X_{t}$
    \State $L_s = CrossEntropy(f_{\theta}(x^{s}), y^{s})$
    \State $L_{t} = CrossEntropy(f_{\theta}(x^{t}), y^{t})$
    \State $L = L_s + L_{t}$
    \State $\theta \gets\underset{\theta}{\mathrm{argmin}} \ L$
\Until{$N_{1}$ Times}

\Repeat
    \Repeat
    \State $\{x^{s}, y^{s}\} \gets X_{s}$
    \State $\{x^{t}, y^{t}\} \gets X_{t}$
    \State \textbf{Algorithm} \ref{alg:a2} $(\{x^{s}, y^{s}\}, \{x^{t}, y^{t}\}, \theta, \phi )$
    \Until{$N_{2}$ Times}
    
    \Repeat
    \State $\{x^{s}, y^{s}\} \gets X_{s}$
    \State $\{x^{t}, y^{t}\} \gets X_{t}$
    \State \textbf{Algorithm} \ref{alg:a3} $(\{x^{s}, y^{s}\}, \{x^{t}, y^{t}\}, \theta, \phi )$
    \Until{$N_{3}$ Times}
    
\Until{$G$ Times}

\EndProcedure
\end{algorithmic}
\end{algorithm}

\begin{algorithm*}
\caption{Meta-Learning of Pixel Weight Model for Source Data}\label{alg:a2}
\begin{algorithmic}[1]
\State \textbf{Input}: $\{x^{s}, y^{s}\}, \{x^{t}, y^{t}\}, \theta, \phi, \alpha, \beta$ 

\Procedure{:}{}

    \State $W(\phi) = f_{\phi}(concat(x^{s}, {y}^{s}))$  \Comment{Compute W and create computation graph for $\phi$.}
    
    \State $L_s = CrossEntropy(f_{\theta}(x^{s}),  y^{s}, W(\phi))$
    \Comment{Cross entropy loss on source domain with pixel weight.}
    
    \State $g_{\theta}(\phi) = \frac{\partial L_{s}}{\partial \theta}$
    \Comment{Computation gradient on $ \theta $ and retain computation graph for $\phi$.}
    
    \State $\theta^{+}(\phi) = \theta - \alpha g_{\theta}(\phi)$
    \Comment{One gradient step on $\theta$.}
    
    \State $L_{t}(\phi) = CrossEntropy(f_{\theta^{+}(\phi)}(x^{t}),  y^{t})$ \Comment{Meta learning loss, the supervision loss on target domain}
    \State $\phi \gets \phi - \beta \frac{\partial L_{t}}{\partial \phi}$ \Comment{Update $\phi$, by gradient-on-gradient}

\EndProcedure

\end{algorithmic}

\end{algorithm*}

\begin{algorithm*}
\caption{Training Segmentation Network with Weighted Source Loss}\label{alg:a3}
\begin{algorithmic}[1]
\State \textbf{Input}: $\{x^{s}, y^{s}\}, \{x^{t}, y^{t}\}, \theta, \phi, \alpha$

\Procedure{:}{}
\State $W = f_{\phi}(concat(x^{s}, {y}^{s}))$ \Comment{Compute W and detach it from computation graph}
    
    \State $L_s = CrossEntropy(f_{\theta}(x^{s}),  y^{s}, W)$  \Comment{Loss on source domain, weighted}
    
    \State $L_{t} = CrossEntropy(f_{\theta}(x^{t}),  y^{t})$  \Comment{Loss on target domain, unweighted}
    
    \State $L = L_s + L_{t}$
    \State $\theta \gets \underset{\theta}{\mathrm{argmin}} \ L$  \Comment{Update the network}

\EndProcedure
\end{algorithmic}
\end{algorithm*}

\section{Experiments}
\subsection{Datasets}
There are three datasets used in this study, GTA5 \cite{Richter_2016_ECCV}, and Synthia are used as the synthetic source datasets and CityScapes \cite{Cordts2016Cityscapes} is the real target dataset.

\textbf{GTA5} has 24966 images generated by the back-end engine of the video game GTA5. Those images are taken from urban scenes of a virtual city that was built upon Los Angeles. There are 19 semantic label classes compatible with the CityScapes dataset and we use the whole dataset as our source synthetic set. 

\textbf{Synthia} is another virtual dataset that contains images from a virtual city. We use SYNTHIA-RAND-CITYSCAPES split with 9400 images as the source synthetic dataset. The label classes in this split are also compatible with the CityScapes dataset.

\textbf{CityScapes} has images of a real-world urban scene taken from multiple European cities. It is a dataset with high resolution ($2048 \times 1024$) and 19 semantic classes. We use the fine-grained part of this dataset which has 2975 images for training (as our real target dataset) and 500 images for validation. Our results are reported on the validation set. Due to the memory constraint on GPU, we follow the prior work \cite{sun2019CVPR} of down-sampling the resolution of images to $1024 \times 512$.

\subsection{Netwrok Architecture}
To make a fair comparison to Hierarchical Region Selection \cite{sun2019CVPR}, for segmentation network $f_{\theta}$, we use the same FCN-8 with VGG-16 backbone. The VGG-16 is pretrained on ImageNet. We empirically find that employing a separate first block of FCN for source and target tasks is beneficial. See figure \ref{fig:arch} for details. 

The weighting network $f_{\phi}$ is a U-Net \cite{10.1007/978-3-319-24574-4_28} whose parameters are randomly initialized. The output of $f_{\phi}$ has a channel of 1, which is similar to $W^{1}$ in Hierarchical Region Selection. The weighting network is only used for training and is discarded at the evaluation. Thus during the testing, the segmentation model we use is the same as Hierarchical Region Selection \cite{sun2019CVPR}.

\subsection{Training Details}
For learning the segmentation network $f_{\theta}$, we follow the hyper-parameters used from Hierarchical Region Selection \cite{sun2019CVPR}. To be specific, Adam optimizer is used with $\beta_{1}=0.9$ and $\beta_{2}=0.999$. The learning rate starts at $1e-4$ and follows a polynomial decay with a power of $0.9$. The images are resized to $1024 \times 512$ and batch size is $1$. For meta learning the weighting model $\phi$, we use another Adam optimizer using the same parameters but without learning rate decay. 

For our method MetaPix, we alternate the training of segmentation network $f_{\theta}$ and weighting network $f_{\phi}$. Each training generation of segmentation network runs for 15K ($N_{3}$ in Algorithm \ref{alg:a1}), and meta training of weighting network runs for 10K ($N_{2}$ in Algorithm \ref{alg:a1}). We run a total of 3 generations ($G$ in Algorithm \ref{alg:a1}).

To get a good initialization for the meta learning, we first pretrain the network $\theta$ with simple direct joint training (unweighted of synthetic data) with synthetic and real data for 150K steps ($N_{1}$ in Algorithm \ref{alg:a1}).

\begin{figure*}[h]
\centering
\includegraphics[width=\textwidth]{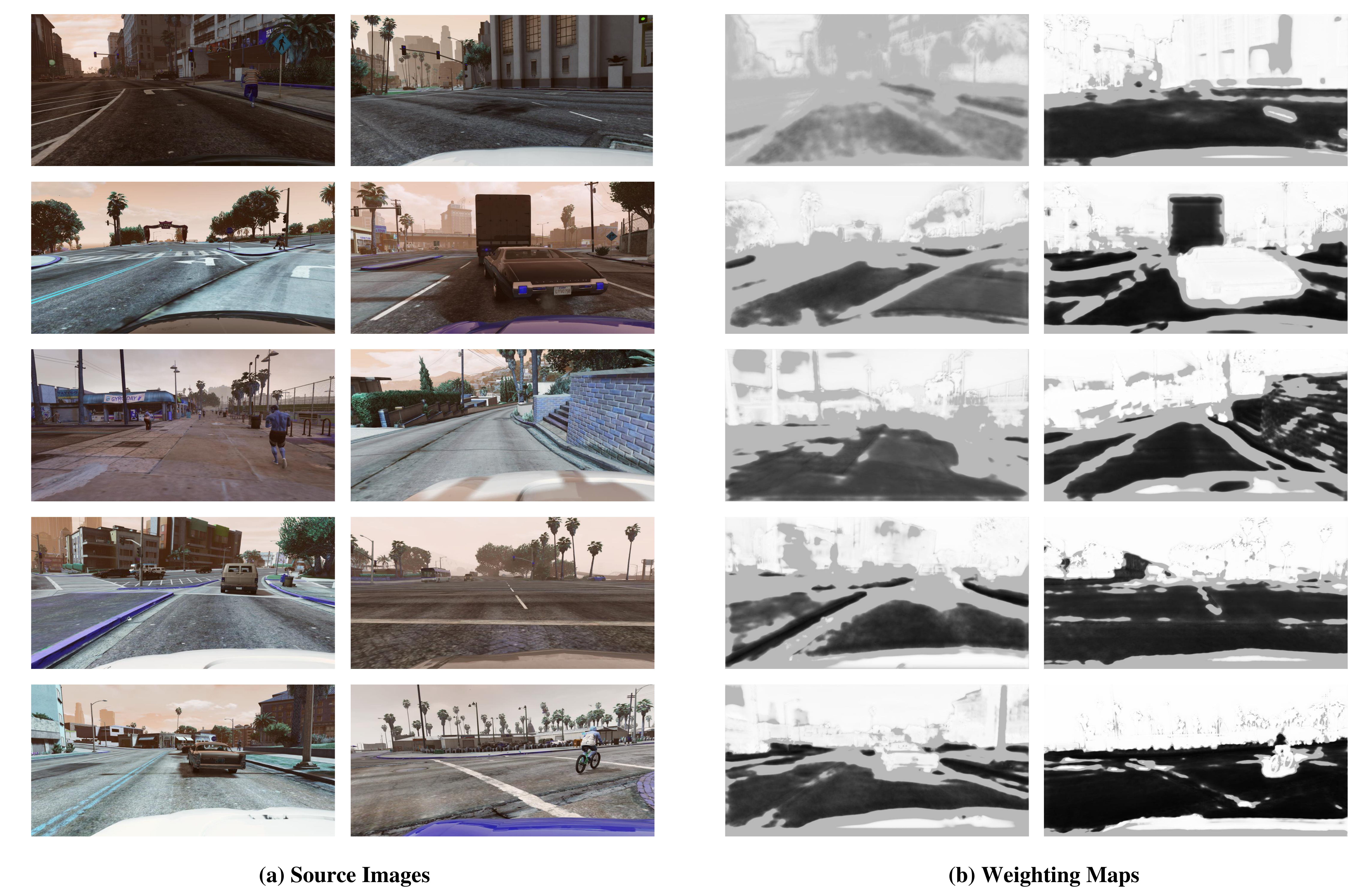}
\caption{Visualization of source images and corresponding weighting maps from GTA5. The brighter part of the weighting map corresponds to the higher weight. Conversely, the darker pixels correspond to lower weight values.}
\label{fig:w}
\end{figure*}

\begin{table}[h]
\begin{center}
\caption{Experimental results of joint learning using GTA5 and CityScapes. $\dagger$ indicates the number that is directly taken from the paper of Hierarchical Region Selection (HRSelect)~\cite{sun2019CVPR}}
\label{table:GTA}
\scalebox{0.8}{
\begin{tabular}{l  c c c c}
		   \hline
		   Method   &  Setting & mIoU \\ \hline
           Swami \etal~\cite{DBLP:conf/cvpr/Sankaranarayanan18}  &  Un- & $37.1\%^{\dagger}$ \\ 
           CL~\cite{DBLP:conf/eccv/ZhuZYSL18}   &  Un- & $38.1\%^{\dagger}$ \\
           ROAD~\cite{DBLP:conf/cvpr/Chen0G18}   &  Un- & $35.9\%^{\dagger}$ \\
           \hline
           FCN    &  - & $65.3\%^{\dagger}$ \\
           PixelDA~\cite{DBLP:conf/cvpr/BousmalisSDEK17}   &  Joint- & $66.1\%^{\dagger}$ \\ 
           FCN+Focal Loss~\cite{DBLP:conf/iccv/LinGGHD17}   & Joint- & $66.2\%^{\dagger}$\\ 
           Direct Joint Training   &  Joint- & $64.6\%^{\dagger}$ \\
           Target Funetuning   &  Joint- & $66.0\%^{\dagger}$ \\
           FCN+GAN    &  Joint- & $64.0\%^{\dagger}$\\ 
           FCN+0-1 Conf. Mask   &  Joint- & $63.7\%^{\dagger}$\\
           FCN+HRSelect-$W^1$~\cite{sun2019CVPR}   &  Joint-  & $66.5\%^{\dagger}$\\
           FCN+HRSelect-$W^1$+$L_{rec}$+$L_{adv}$~\cite{sun2019CVPR}   &  Joint- & $67.6\%^{\dagger}$\\
           \hline
           FCN (our implementation)    &  - & $64.7\%$ \\
           FCN+FDA~\cite{DBLP:conf/cvpr/0001S20}   &  Joint- & 66.5\% \\
           FCN+MinEnt~\cite{vu2018advent}   &  Joint- & 66.3\% \\
           FCN+MetaPix-$W^{1}$ (ours)   &  Joint- & $\mathbf{68.7\%}$ &
           \\ \hline

\end{tabular}
}
\end{center}
\vspace{-0.3in}
\end{table}

\subsection{Results on GTA5 $\xrightarrow[]{}$ CityScapes}
In this section, we provide quantitative results on using GTA5 as a source synthetic dataset and CityScapes as a target real dataset. 

We compare directly to Hierarchical Region Selection \cite{sun2019CVPR}. Because we discard the weighting network during testing, the segmentation model in testing we use is the same as Hierarchical Region Selection. To verify the effectiveness of using meta learning for the pixel-weighting network, it is fairest to compare to FCN+HRSelect-$W^{1}$ in Hierarchical Region Selection. 

Our method MetaPix with mIOU of 68.7\% outperforms 66.5\% of FCN+HRSelect-$W^{1}$ by a large margin of 2.2\%. Our method does not include any additional tricks commonly used in UDA (because we mainly intend to justify the use of meta learning to learn the weighting maps). But still, MetaPix outperforms the hybrid method FCN+HRSelect-$W^{1}$+$L_{rec}+L_{adv}$ with a reconstruction loss and an adversarial loss on feature space by 1.1\%. Considering that the baseline method (training with the target set only) has mIOU of 65.3\%, our improvement over Hierarchical Region Selection is significant.

We also notice that directly adapting the techniques from UDA to Domain Transfer Setting will not work well. Besides the PixelDA~\cite{DBLP:conf/cvpr/BousmalisSDEK17} and FCN+GAN shown in the previous work \cite{sun2019CVPR}, we also try with two latest UDA methods, \ie, FDA~\cite{DBLP:conf/cvpr/0001S20} and MinEnt~\cite{vu2018advent}. FDA uses Fourier Transform to minimize the domain gap at the image level and MinEnt minimizes the entropy at the feature level. We add the supervision loss from the labeled real target dataset to both FDA and MinEnt. None of these methods can substantially improve over the baseline.

\begin{table}[h]
\begin{center}
\caption{Experimental results of joint learning using Synthia and CityScapes. $\dagger$ indicates the number that is directly taken from the paper of Hierarchical Region Selection (HRSelect)~\cite{sun2019CVPR}}
\label{table:Synthia}
\scalebox{0.8}{
\begin{tabular}{l  c c c c}
		   \hline
		   Method   & Setting & Mean IoU \\ \hline
           Swami \etal~\cite{DBLP:conf/cvpr/Sankaranarayanan18}  &  Un- & $34.8\%^{\dagger}$ \\ 
           CL~\cite{DBLP:conf/eccv/ZhuZYSL18}   &  Un- & $34.2\%^{\dagger}$ \\
           ROAD~\cite{DBLP:conf/cvpr/Chen0G18}   &  Un- & $36.2\%^{\dagger}$ \\
           \hline
           FCN    &  - & $65.3\%^{\dagger}$ \\ 
           PixelDA~\cite{DBLP:conf/cvpr/BousmalisSDEK17}   &  Joint- & $64.0\%^{\dagger}$ \\
           Direct Joint Training   &  Joint- & $62.9\%^{\dagger}$ \\
           Target Funetuning   &  Joint- & $64.8\%^{\dagger}$ \\
           FCN+GAN    &  Joint- & $62.6\%^{\dagger}$\\ 
           FCN+HRSelect-$W^1$+$L_{rec}$+$L_{adv}$~\cite{sun2019CVPR}  &  Joint- & $66.3\%^{\dagger}$\\
           \hline
           FCN (our implementation)    &  - & $64.7\%$ \\
           FCN+FDA~\cite{DBLP:conf/cvpr/0001S20}   &  Joint- & 63.6\%  \\
           FCN+MinEnt~\cite{vu2018advent}   &  Joint- & 65.4\% \\
           FCN+MetaPix-$W^{1}$ (ours)   &  Joint- & $\mathbf{66.7\%}$ &
           \\ \hline

\end{tabular}
}
\end{center}
\vspace{-0.3in}
\end{table}

\subsection{Results on Synthia $\xrightarrow[]{}$ CityScapes}
In this section, we test our method using Synthia as source synthetic dataset and CityScapes as target real dataset. The original results from Hierarchical Region Selection \cite{sun2019CVPR} do not include FCN+$W^{1}$. Thus, we compare MetaPix to the hybrid method with reconstruction loss and adversarial training in Hierarchical Region Selection. This may put our method at a disadvantage, as those common tricks are known to improve performance. However, still, our method outperforms the hybrid method (66.3\%) by 0.4\%.

Similar to what we find in the previous section, add a supervision loss of labeled target dataset to UDA methods like FDA or MinEnt will not improve over the baseline.

\subsection{Ablations}
\label{sec:AB}

\begin{figure}[h]
\centering
\includegraphics[width=0.45\textwidth]{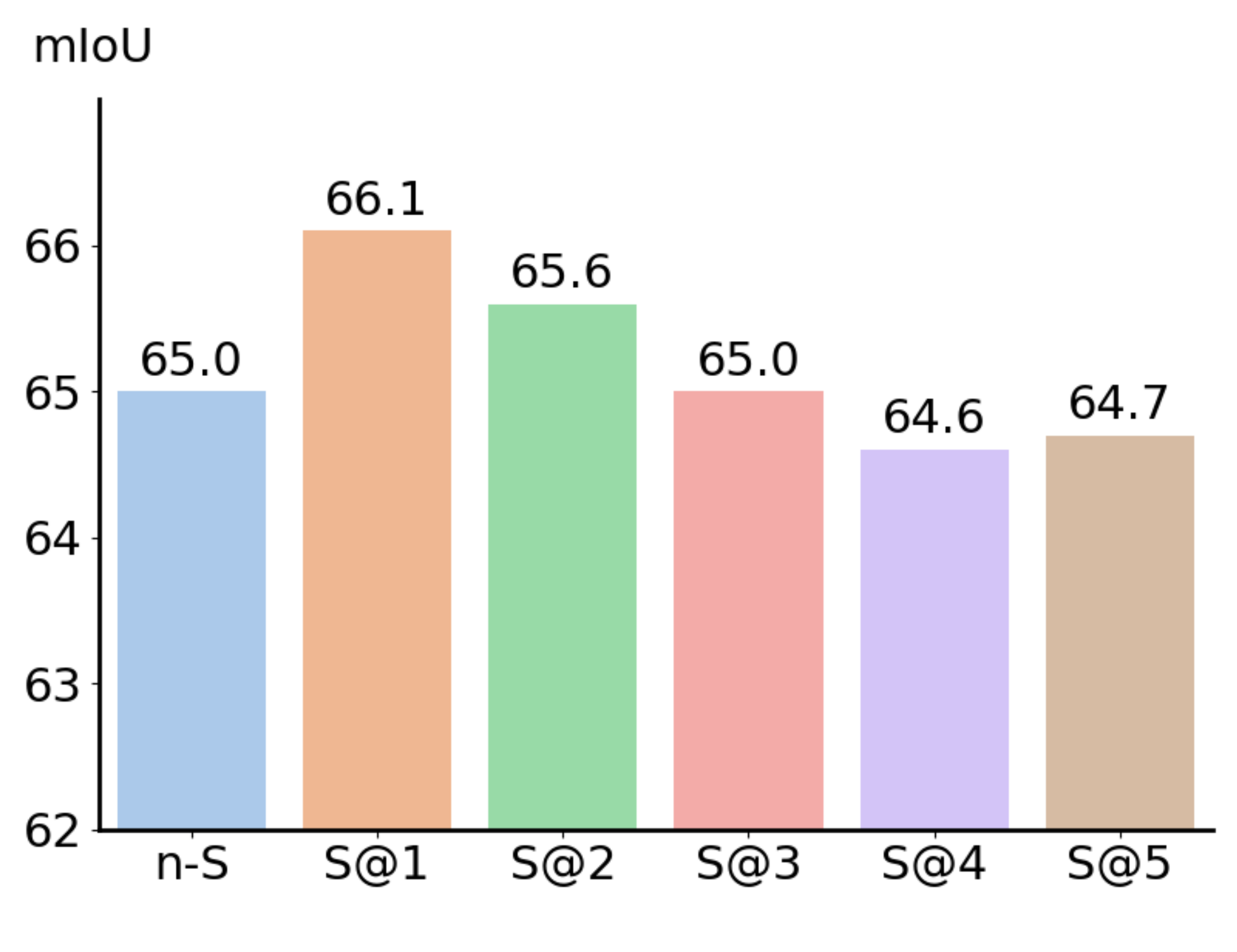}
\caption{Joint training of GTA5 and CityScapes with different architectures. The model splits at different blocks with the specific heads for GTA5 and CityScapes. For example, S@1 denotes splitting the network at the first block of segmentation network (See Figure \ref{fig:arch} (c) for visualization).}
\label{fig:split}
\end{figure}

As discussed in the section of network architecture, we find it beneficial to split the segmentation network at the first block for source and target task, \ie, the gradient from the source task will not update the first block of the segmentation network. We show the results of direct joint training (baseline method) of source and target data with (1) no splitting, (2) split at the 1st block, (3) splitting at the 2nd block, (4) splitting at the 3rd block, (5) splitting at the 4th block (6) splitting at the 5th block. Note that "no split" is the same as "direct joint" from Hierarchical Region Selection and "splitting at the 5th block" (no sharing parameters for source and target task) is equal to "FCN" target only from Hierarchical Region Selection. As shown in Figure \ref{fig:split}, we get an improvement of 1.1\% over sharing the whole network for both source and target tasks.

\begin{table}[h]
\begin{center}
\caption{Experiments of MetaPix on GTA5 $\xrightarrow[]{}$ CityScapes, with different weighting models: w/o $W$, w/ $W^{19}$ and w/ $W^{1}$.}
\label{table:W1 or W19}
\scalebox{0.8}{
\begin{tabular}{l l c }
           \hline
           Source $\xrightarrow[]{}$ Target & Method   &  Mean IoU \\
           \hline
           GTA5 $\xrightarrow[]{}$ CityScapes & FCN+finetune 45K steps       &  $66.4\%$  \\
           GTA5 $\xrightarrow[]{}$ CityScapes & FCN+MetaPix-$W^{19}$ (ours)  &  $68.3\%$ \\
           GTA5 $\xrightarrow[]{}$ CityScapes & FCN+MetaPix-$W^{1}$ (ours)   &  $\mathbf{68.7\%}$ 
           \\ \hline

\end{tabular}
}
\end{center}
\vspace{-0.2in}
\end{table}

In Section 3.1, our weighting network outputs a tensor of size $[B, 1, H, W]$ which we denote as $W^{1}$ to weight the cross-entropy loss at each pixel. Another design choice is to output a  tensor of size $[B, 19, H, W]$ (which we denote as $W^{19}$), where 19 is the number of classes in the source dataset, \ie, GTA5 and Synthia. The weighting of $W^{1}$ will treat each channel of the output equally, whereas the weight of $W^{19}$ computes different weights for each channel. Our experiment in Table \ref{table:W1 or W19} shows that $W^{1}$ works slightly better than $W^{19}$.

The total number of training steps for the segmentation network is $195K$, which has $150K$ for joint training followed by three generations of further training with adaptive meta weighting $W^{1}$, each with $15K$ steps. In the first row of Table \ref{table:W1 or W19}, we show the experiment with 150K steps of joint training in GTA5 and CityScapes followed by further training of 45K steps without the meta learned weighting model. "FCN+finetune 45K steps" with 66.4\% indicates that further training the network without meta weighting models will not contribute to a significant improvement of accuracy.

\section{Conclusion}
In this paper, we propose MetaPix, a meta learning method for learning an adaptive weight map for source task in domain transfer learning of semantic segmentation. MetaPix has only one single module to be meta learned in an end-to-end manner. The training procedure is simple compared with most hybrid UDA methods that have multiple modules to be trained by GAN. The experimental results of MetaPix prove the effectiveness of using meta learning to learn weighting maps over the heuristic to select source pixels that are similar to target data proposed in a latest work.

{\small
\bibliographystyle{ieee_fullname}
\bibliography{egbib}
}

\end{document}